\documentclass[letter, 12pt, onecolumn]{ieeetran}
\IEEEoverridecommandlockouts
\let\labelindent\relax
\usepackage{graphicx}
\usepackage{eqlist}
\usepackage{amsfonts}
\usepackage{tabularx}
\usepackage{booktabs}
\usepackage{amssymb}
\usepackage{amsmath}
\usepackage{enumitem}
\usepackage{threeparttable}
\usepackage[lined, ruled, linesnumbered, commentsnumbered]{algorithm2e}
\usepackage{lipsum}

\usepackage[usenames,dvipsnames]{xcolor}
\usepackage{comment}
\usepackage{flushend}
\usepackage{todonotes}
\usepackage{xcolor}
\usepackage{booktabs}

\begin{document}
\title{Planning to Repose Long and Heavy Objects Considering a Combination of Regrasp and Constrained Drooping}

\author{Mohamed Raessa$^{1}$, Weiwei Wan$^{\ast1}$,
Keisuke Koyama$^1$, and Kensuke Harada$^{12}$
\thanks{$^\ast$ Weiwei Wan is the corresponding author.
Email: wan@sys.es.osaka-u.ac.jp.
$^1$ Graduate School of Engineering Science, Osaka University, Osaka,
Japan. $^2$ National Institute of Advanced Industrial Science and Technology (AIST), Tsukuba, Japan.}}
%\author{\textit{Author names are hidden for blind review}}
\maketitle

\begin{abstract}
\noindent\textbf{Purpose of this paper:}
This paper presents a hierarchical motion planner for planning the manipulation motion to repose long and heavy objects considering external support surfaces.\\
\textbf{Design/methodology/approach:}
The planner includes a task level layer and a motion level layer. We formulate the manipulation planning problem at the task level by considering grasp poses as nodes and object poses for edges. We consider regrasping and constrained in-hand slip (drooping) during building graphs and find mixed regrasping and drooping sequences by searching the graph. The generated sequences autonomously divide the object weight between the arm and the support surface and avoid configuration obstacles. Cartesian planning is used at the robot motion level to generate motions between adjacent critical grasp poses of the sequence found by the task level layer. \\
\textbf{Findings:}
Various experiments are carried out to examine the performance of the proposed planner. The results show improved capability of robot arms to manipulate long and heavy objects using the proposed planner.\\
\textbf{What~is~original/value~of~paper:}
Our contribution is we initially develop a graph-based planning system that reasons both in-hand and regrasp manipulation motion considering external supports. On one hand, the planner integrates regrasping and drooping to realize in-hand manipulation with external support. On the other hand, it switches states by releasing and regrasping objects when the object is in stably placed. The search graphs' nodes could be retrieved from remote cloud servers that provide a large amount of pre-annotated data to implement cyber intelligence.
\end{abstract}

\section{Introduction}
Manipulating long and heavy objects using a single robot arm is challenging because of robots and grippers' limited duty. This difficulty originates from the objects' shapes and masses. They dictate how external forces, such as gravity and inertia, affect the object's stability during the manipulation process. Previously, researchers considered overcoming the problem by using multiple robots to share object weight. The examples include but not limit to using multiple arms \cite{sina2016coordinated,xian2017closed, dehio2018modeling}, multiple mobile robots \cite{yamashita1998cooperative}, multiple mobile manipulators \cite{alipour2011point}, or a combination of robots and other machines \cite{ohashi2016realization}. The drawback is using multiple robots decreases the overall automation system's efficiency because of the high costs. Also, the complications associated with multi-robot motion planning adds additional overhead to the system. To reduce the costs, this paper proposed to plan manipulating heavy objects using a single arm while keeping it partially supported by a support surface. 

We consider regrasping and constrained drooping for effective maneuvering and in-hand pose adjustment. Especially, drooping refers to the in-hand sliding motion caused by gravitational torque. The earliest studies worked on drooping manipulation are \cite{aiyama1993pivoting,carlisle1994pivoting}. In our previous research \cite{raessa2020human}, we examined the reasons behind the drooping motion associated with heavy objects manipulation, and implemented a constrained motion planner to eliminate it. In this paper we take advantage of our understanding about the drooping to transit grasp poses and realize in-hand pose adjustment. We consider constraining the drooping motion by moving the robot gripper's height above a support surface in a limited. One end of the object is grasped throughout a task while the other end is kept in contact with the support surface. The heavy object weight is shared between the gripper and the support surface. Meanwhile, the holding gripper's height is adjusted in a range computed considering the object's shape and the difference between the gripper's current pose and a goal in-hand pose. We formulate the manipulation planning problem by considering grasp poses as nodes and object poses for edges. We use hierarchical motion planning approaches to autonomously determine regrasp and drooping sequences and generate robotic manipulation motion. 

Our development is based on several assumptions about the difficulties as follows.
\begin{enumerate}
    \item The grasped object is long and heavy enough to slip and rotate in a parallel robot gripper. We refer to the slippage--rotation motion in the parallel gripper drooping.
    \item The object needs to remain in contact with the support surface during manipulation. The surface fully supports the object's weight while being regrasped and partially support it during drooping.
    \item The gripper finger pads are made of soft materials, which enables the gripper to exert friction torque on the object while partially holding it. The soft finger contact assumption allows dividing the object's weight between the gripper and the support surface.
\end{enumerate}

\begin{figure}[!htbp]
    \centering
    \includegraphics[width=.33\columnwidth]{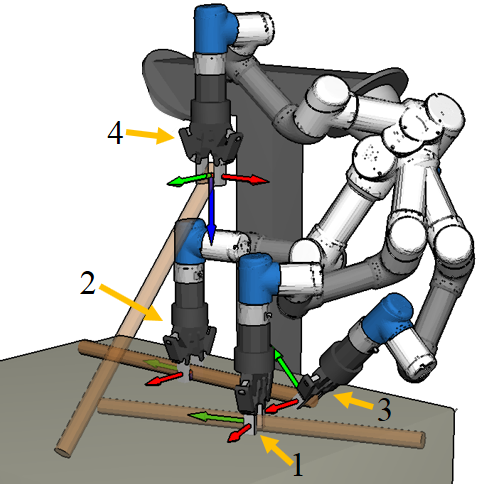}
    \caption{An example of the robot motion sequence found by our planner. The robot conducts regrasp from ID (1) to (3), and performs constrained drooping considering the table as a support from ID (3) to (4). The object is successfully delivered to a goal pose at ID (4) with the help combined regrasping and drooping.}
     \label{fig:regroop-integrated}
\end{figure}

We model and develop a combined regrasp and drooping planner based on these assumptions and examine our development using real-world experiments. The results show that our method can successfully find manipulation sequences for a robot to maneuver long and heavy objects. The robot may autonomously determine the switches between grasping poses and in-hand drooping poses and finish reposing tasks. Fig.\ref{fig:regroop-integrated} shows an example of the robot motion sequence found by our planner. Here, the robot cannot fully lift the stick. Given the start and goal poses, our planner finds a sequence (ID (1)-(4) in the figures) to deliver the stick to the goal pose at 4.

This paper is organized as follow, Section \ref{sec-related} presents related work. Section \ref{sec-plan-drooping} explains the drooping manipulation planning and grasp transition criteria. Section \ref{sec-plan-regrasp} describes the regrasp planning and the integration of drooping and regrasp. Section \ref{sec-exp} presents the experiments and analysis. Section \ref{sec-dis} carries out a discussion about the proposed system performance and highlights its good points and challenges. Section \ref{sec-conclude} concludes the study and discusses the potential future directions of improvements.

\section{Related Work}
\label{sec-related}
\subsection{Heavy Objects Manipulation}
Different methods have been developed to solve the problem of heavy object manipulation. Those methods assume that a single robot is not enough to get such tasks done and employ multiple robots for help. For example, a method for coordinating a multi-arm system's motion to receive an object from a human handover was proposed in \cite{sina2016coordinated}. In \cite{xian2017closed}, the authors explored the changes in the configuration space connectivity when the multi-arm system work together in a closed chain to manipulate a large object. A method for stable planning of heavy object carrying tasks using mobile manipulators was presented in \cite{alipour2011point}. The authors formulated the motion planning problem for several robots as an optimization problem with a cost function that minimizes the mobile base motion. In \cite{moosavian2010heavy}, a hybrid system consisting of a serial manipulator attached to a mobile Stewart mechanism was proposed. The aim was to provide stable maneuvers through the analysis of postural stability of the combined system components. An approach that uses a group of mobile robots with a handcart for heavy objects transportation was presented in \cite{ohashi2016realization}. In \cite{dehio2018modeling}, the authors used four robot arms to manipulate heavy objects. The object was modeled as a virtual link to include in the dynamic model to improve the task accuracy.

Our study proposes an approach that enables a single robot arm to manipulate heavy objects with the help of a support surface in the arm's workspace. The gravitational torque that long and heavy objects experience while being manipulated is carefully controlled to adjust the object's in-hand pose.

\subsection{Manipulation with Regrasping}
Industrial manipulators usually use simple two-jaw grippers to interact with the environment. Such grippers do not possess enough dexterity required for manipulation tasks \cite{bicchi2000hands}. Therefore, multiple methods such as regrasping \cite{tournassoud1987regrasping}, vision-based grasping \cite{liu2014vision}, and dual-arm manipulation \cite{ambrose2000robonaut} have been developed to fulfill the need for dexterity. In this study, the first manipulation primitive motion we use is regrasping. In regrasping, the existence of an external surface within the manipulator workspace is assumed. The surface makes it possible to obtain stable placements of the manipulated objects. The manipulation process becomes a search for a sequence of stable placements of the object that connect the object's start pose to the end pose. The transition between the different grasps in this sequence is made by breaking the grasp and moving to another grasp at the same object's stable placement pose. In our previous work \cite{wan2015reorientating, wan2017regrasp}, we implemented regrasping through graph search in three different steps -- grasp planning, placement planning, and graph construction. Then, we performed the regrasp task planning by searching the shortest path between the start and end poses of the object \cite{wan2017regrasp,calandra2018more}.

In this study, we integrate regrasping and constrained drooping to extend a robot arms' manipulation capability. A robot can manipulate objects with autonomously determined regrasp and drooping considering minimum times of adjustment. The regrasp is used for discrete in-hand pose adjustment. The constrained drooping is used for continuous adjustment without releasing.

\subsection{Manipulation with External Forces}
During constrained drooping, one end of the grasped heavy object is allowed to slip in-hand under the effect of gravity. Meanwhile, the other end is kept in contact with the support surface in a controlled way to maintain the desired grasps or transit between them. From a broader view, the constrained drooping is one example of ``manipulation with external forces and contacts'', namely extrinsic manipulation. The gravitational torque is the external force that induces the change of the in-hand pose. The in-hand slip is limited by keeping the other end of the object always in contact with an external surface. Previous research that presented multiple ways to manipulate objects with external contacts and a simple gripper is available in \cite{dafle2014extrinsic}. Multiple non-prehensile approaches for object manipulation were also implemented. Examples include but not limit to planar pushing \cite{hogan2020feedback, lynch1996stable, zhou2016convex}, pushing against external supports \cite{chavan2015prehensile, holladay2015general, hou2018fast}, pivoting \cite{sawasaki1991tumbling,carlisle1994pivoting}. 

\section{Constrained Drooping and Grasp Transition Criterion}
\label{sec-plan-drooping}
In our research, we consider reposing manipulation using drooping or in-hand slip caused by gravitational toque. The following sub-sections explain the principles of how a gravitational torque induces the drooping motion and how we use it for grasp reposing manipulation. Especially, we focus on constrained drooping, where a support surface keeps up one end of the grasped heavy object while the object body slips and rotates in-hand under the influence of gravity. We plan the robot motion to ensure the other end continuously contacts with the table surface in a controlled way to maintain the desired grasp poses or transit between them.

\subsection{Gravity Torque Effect}
\label{subsection:gravitytorque}
When a two-finger parallel gripper manipulates long and heavy objects, they become prone to slippage in-hand (or drooping) due to the effect of gravity torque. The gravitational torque determines the drooping motion and the gripper's frictional torque \cite{raessa2020human}. When a parallel gripper gets inclined, the gravitational torque around the jaw opening direction increases. A larger inclination would further increase the gravitational torque, and at a certain instant, the gravitational torque may exceed the maximum friction torque of the gripper's finger pads and causes the grasped object to droop in the robot hand. The following equation relates the gravity torque to the various parameters that affect drooping.
\begin{gather}
    T_g = \dfrac{mg}{2} sin(\theta) sin(\phi) (Obj_{CoM_{rel-EE}}  \hspace{1mm} sin(\phi)) \nonumber \\
    + \dfrac{mg}{2} sin(\theta) cos(\phi) (EE_{length}+Obj_{CoM_{rel-EE}} \hspace{1mm} cos(\phi)), 
\raisetag{2.5\normalbaselineskip}
\label{eq1}
\end{gather}

By observing  Equation \eqref{eq1} we understand that the parameter with the most significant influence on the gravity torque is the inclination angle $\theta$. Thus, in our drooping-based manipulation approach, we maximize an object's drooping by keeping the inclination angle $\theta$ at its maximum during the manipulation task. In the next subsection, we explain how we use drooping to realize in-hand pose adjustment and reach to grasp transitions.

\subsection{Grasp Transition Criterion}
\label{subsec-trans}
Based on Equation\eqref{eq1}, we operate the robot within the range of a gripper inclination angle that causes the maximum possible drooping motion. An object can freely droop in-hand within this range while being partially grasped by the gripper and kept up by a support surface. The support surface acts as an external pusher and changes the object's in-hand pose as the gripper moves upward or downward. We name such a change constrained drooping. In our proposed planner, we employ constrained drooping to change the in-hand grasp poses. By properly sequencing the gripper's upward and downward motion, a robot can autonomously change grasp poses and hence object poses. Essentially, the criterion of grasp transition depends on changing the gripper's height over the support surface with a distance equivalent to the change in angle between two consecutive grasps. This criterion is described by Equations \eqref{eq2} and \eqref{eq3} for the upward and downward motions.
\begin{equation}
\label{eq2}
    d_{up} = l_{stick} [sin(\theta_{stick_{init}} + \theta_{target_{grasp}}) - sin(\theta_{stick_{init}})]
\end{equation}
\begin{equation}
\label{eq3}
    d_{down} = l_{stick} [sin(\theta_{stick_{init}} - sin(\theta_{stick_{init}} - \theta_{target_{grasp}})]
\end{equation}

\begin{figure}[!htbp]
    \centering
    \includegraphics[width=.67\columnwidth]{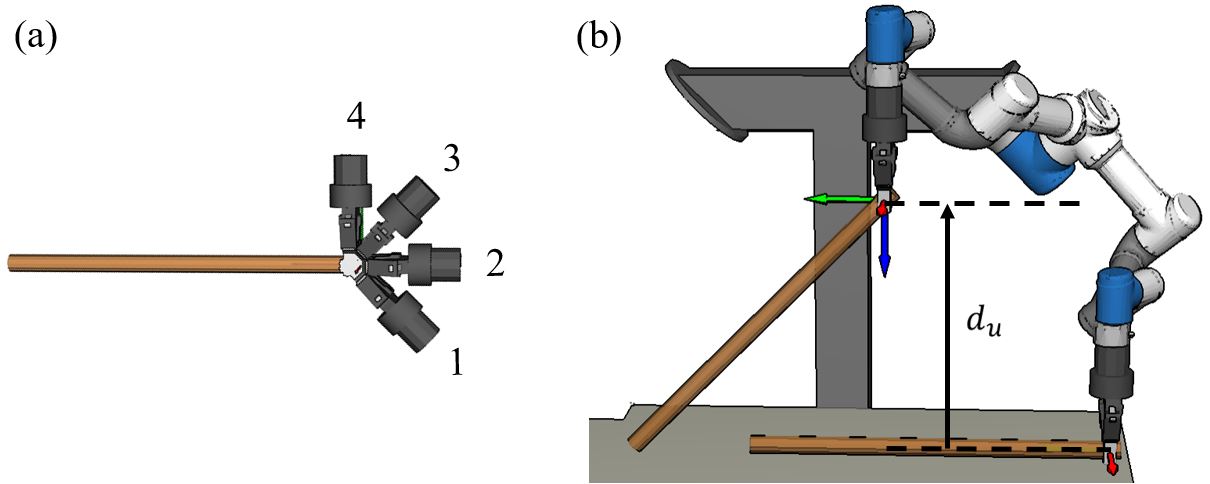}
    \caption{Drooping based grasp transition for in-hand manipulation. The change of the gripper's height above the support table enables grasp transition between different grasp poses.}
     \label{fig:grasptrans}
\end{figure}

Fig.\ref{fig:grasptrans}(a) shows a set of predefined grasp poses that hold the end of an object. Fig.\ref{fig:grasptrans}(b) illustrates an example of the in-hand pose change based on the drooping motion from grasp pose with ID (4) to grasp pose with ID (3). This transition requires the object to move up a distance equivalent to the angle between the two grasps, which is $45^{\circ}$ in the shown example. The same criterion generalizes to any desired change of grasp angle. This method is also flexible to different horizontal positions as in the manipulation process, the height change defines a plane parallel to the support surface, and any point in the parallel plan fulfills the transition condition according to Equations \eqref{eq2} and \eqref{eq3}. Having a plan that satisfies the condition allows not only grasp transitions but also changing the object translation and orientation at the same time. On the other side, to implement a grasp transition in the other direction, i.e., transit from the grasp pose at ID (3) to (4), the gripper needs to move downward. The conversion criterion between up-down motion and constrained drooping is effective as long as the gripper inclination angle is larger than the drooping threshold.

\section{Hierarchical Planning Considering Regrasping and Constrained Drooping}
\label{sec-plan-regrasp}
We use a hierarchical planning scheme for finding a sequence of constrained drooping grasp poses that change the pose of an object. At the task level, we employ a graph-based planner to generate sequences of object poses between the start and the goal pose. We build a graph of grasp poses and object poses that satisfy the contact condition and traverse the graph to find a sequence of the object's critical poses and a sequence of grasp poses for manipulating the object between the given start and goal. Each grasp pose is modeled as a node of the graph. They are connected by edges defined considering object poses, robot payload, and the grasp transitions criterion shown in Equations \eqref{eq2} and \eqref{eq3}. 

Besides drooping, we expand our graph with regrasping by connecting the grasp poses associated with stable object poses at the task level. These poses indicate the critical poses for regrasping. By connecting them, we can search across both regrasping and constrained drooping and implemented combined sequence planning.  

At the motion level, we use Cartesian planning to generate robot motions that move the object between the critical poses designated by the task-level planner. The critical grasp poses sequence found at the task level are connected through Cartesian motion planning. Cartesian planning is used because it helps find robot motion trajectories that satisfy the condition of maintaining the contact between the object and the support surface. The following three subsections present the details of the task level planning (A, B) and the Cartesian motion planning (C), respectively.

\subsection{Task Level Planning}
\subsubsection{Drooping manipulation graph}
\begin{figure}[!htbp]
    \centering
    \includegraphics[width=.9\columnwidth]{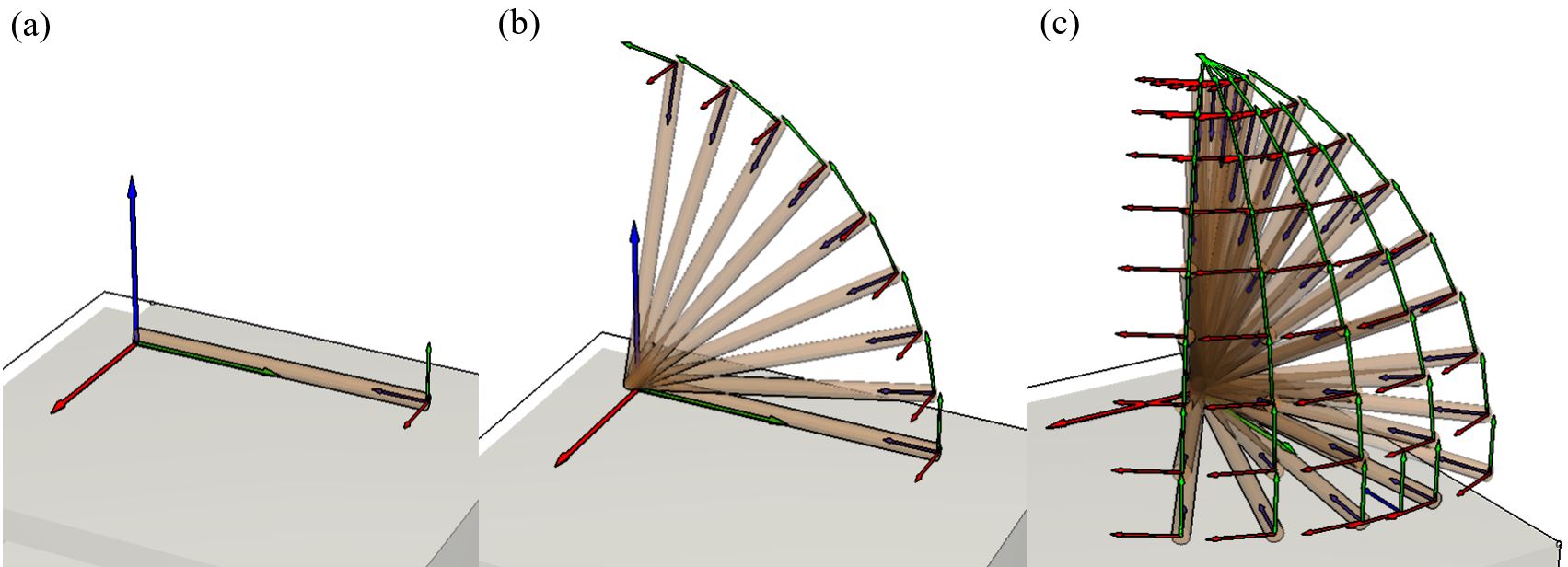}
    \caption{The process of generating object poses with the condition of always being in contact with the support surface. (a) The process starts with a virtual object placed on the support surface. (b) The object is rotated around the $X$ axis of the placement point in steps between $0^{\circ}$ and $90^{\circ}$. (c) Every resulting pose from the previous step is rotated about the $Z$ axis of the placement point and the result is a bouquet of object poses that are sharing the same placement point.}
     \label{fig:make_bouquet}
\end{figure}

The essential requirement for drooping manipulation of heavy objects is to have the object always contact the support surface. This requirement is taken into consideration when designing the graph nodes of a drooping manipulation graph. The process of sampling graph nodes is illustrated in Fig. \ref{fig:make_bouquet}. The process starts with an object at a placement point on the support surface. Starting from this pose, the object is virtually rotated about the $X$ axis of the placement point as shown in Fig. \ref{fig:make_bouquet}(b) to generate many different poses. In the following step, every generated object pose from the previous rotation is further rotated about the $Z$ axis of the placement point as shown in Fig. \ref{fig:make_bouquet}(c). The second set of rotations result in a bouquet of object poses that share the same placement point. All the object poses in a single bouquet satisfy the condition that the object must contact the support surface. After that, we discretize the support surface into a grid of placement points and repeat the bouquet generation process at each point to get the evenly sampled object poses on the whole surface. Then, we transform pre-annotated grasp poses to each of the evenly sampled object poses and filter out the IK reachable and collision-free ones. The remaining grasp poses after filtering are used as the graph nodes. 

After sampling the graph nodes, we connect them to finish up the drooping manipulation graph. Whether the nodes can be connected is determined considering the object poses, robot payload, and the grasp transitions criterion shown in Equations \eqref{eq2} and \eqref{eq3}. The graph is then ready to be searched for finding drooping manipulation sequences after the edges are connected.

\begin{figure}[!htbp]
    \centering
    \includegraphics[width=.97\columnwidth]{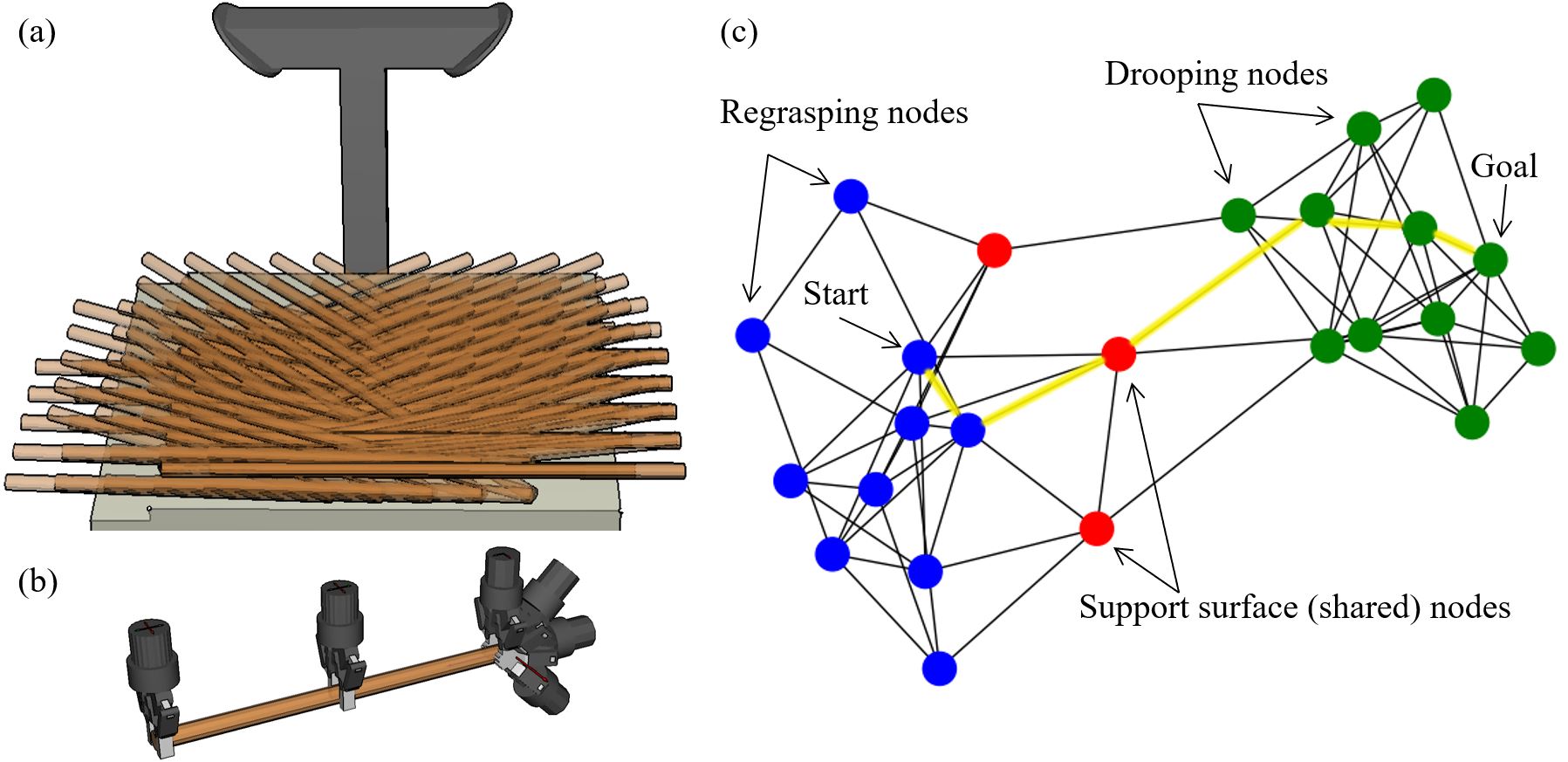}
    \caption{(a) Samples of stable object poses on a support surface. (b) The pre-annotated grasp poses for grasping the stick. The object poses that exist both in (a) and the set of the bouquets in Fig.\ref{fig:make_bouquet}(c) represent the connecting poses. Their associated grasp poses are the candidate connecting nodes between the regrasp graph component and the drooping graph component.}
     \label{fig:table_poses_w_grasps}
\end{figure}

In detail, the edge connection between the graph nodes is determined considering the criterion shown in Equations \eqref{eq2} and \eqref{eq3}. If the height between a pair of consecutive grasps matches the connection criterion, an edge will be established in the graph. This edge is referred to as a grasp transition edge, and it implies a constrained drooping action. Another criterion considered for making edges is to connect graph nodes that share the same grasp pose. This edge is referred to as the translation edge. Such edges allow manipulating objects while maintaining the same relative grasp between the gripper and the object. In this way, the resulted connected graph enables both grasp transitions and object pose translation while being in contact with the support surface.

\subsubsection{Expansion with Regrasping Nodes}
For the advanced dexterity of industrial robots, we may further expand the graph with regrasp nodes. Regrasp nodes and edges consider the stable placement poses on the support surface during a manipulation task. In a regrasp sequence, a robot will release and regrasp objects while they are resting stably on the support surface. Thus, we further sample stable placements and find their associated grasp poses, and connect these nodes to the previously built drooping manipulation graph. Similar to the previous step, the reachable, collision-free grasp poses are included as graph nodes, and the unsatisfactory grasp poses are discarded. Fig. \ref{fig:table_poses_w_grasps}(a) shows an example of the stable placements of a stick on a table surface. Fig. \ref{fig:table_poses_w_grasps}(b) shows the pre-annotated grasp poses. They are associate with each of the sampled object pose to create candidate expansion nodes. The set of object poses that exist in both of the support surface poses and the bouquet poses represent the possible connecting poses between drooping and regrasp. The grasp poses associated with these connecting poses are the shared nodes. They represent the candidate interchange node for switching between drooping and regrasp. Fig. \ref{fig:table_poses_w_grasps}(c) shows an example of the expanded manipulation graph. Here, the regrasping nodes are illustrated in blue. The drooping nodes are illustrated in green. The shared connecting nodes are shown in red. A planned path between a given regrasping start node to a drooping goal node is shown on the graph with yellow color. The expanded graph enables a robot to manipulate objects from any given pose on the support surface into its workspace and complete meaningful tasks.

\section{Experiments and Analysis}
\label{sec-exp}
\begin{figure}[!htbp]
    \centering
    \includegraphics[width=.4\columnwidth]{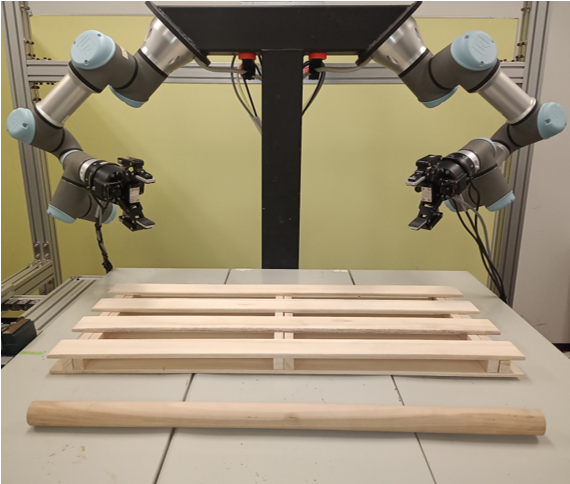}
    \caption{The experimental setup of our system. One of the two UR3 arms and the Robotiq 2F-85 gripper attached to its end flange is used to examine our planner. The objects used in the experiments are shown in front of the robot. They include a wooden stick and a wooden duck-board.}
     \label{fig:exp_setup}
\end{figure}

The experimental setup of our research is shown in Fig. \ref{fig:exp_setup}. We use one of the two UR3 arms and the Robotiq 2F-85 gripper attached to its end flange for object manipulation. The finger pads of the grippers have rubber pads and form soft-finger contacts during grasping. Two wooden objects are prepared to verify the proposed approach's efficacy, including a stick and a duck-board. The various parameters of the objects are listed in Table. \ref{table1}. 

\begin{table}[!htbp]
\centering
\begin{tabular}{@{}ccccc@{}}
\toprule
Object     & Length(mm) & Width(mm) & \begin{tabular}[c]{@{}c@{}}Thickness / \\ Diameter(mm)\end{tabular} & Weight(g) \\ \midrule
Duck-board & 750 & 330 & 35 & 920 \\ \midrule
Stick & 656 & - & 32 & 280        \\ \bottomrule
\end{tabular}
\caption{Dimensions and weights of the objects used in the experiments.}
\label{table1}
\end{table}

We designed two sets of experiments to examine the developed planner. In the first set, we only consider drooping manipulation. The goal is to verify that our method can successfully carry out grasp transitions using the criterion shown in section \ref{subsec-trans}. The second set concentrates on the hierarchical planner's efficacy in generating motion sequences of combined regrasping and drooping.

\begin{figure}[t]
    \centering
    \includegraphics[width=\columnwidth]{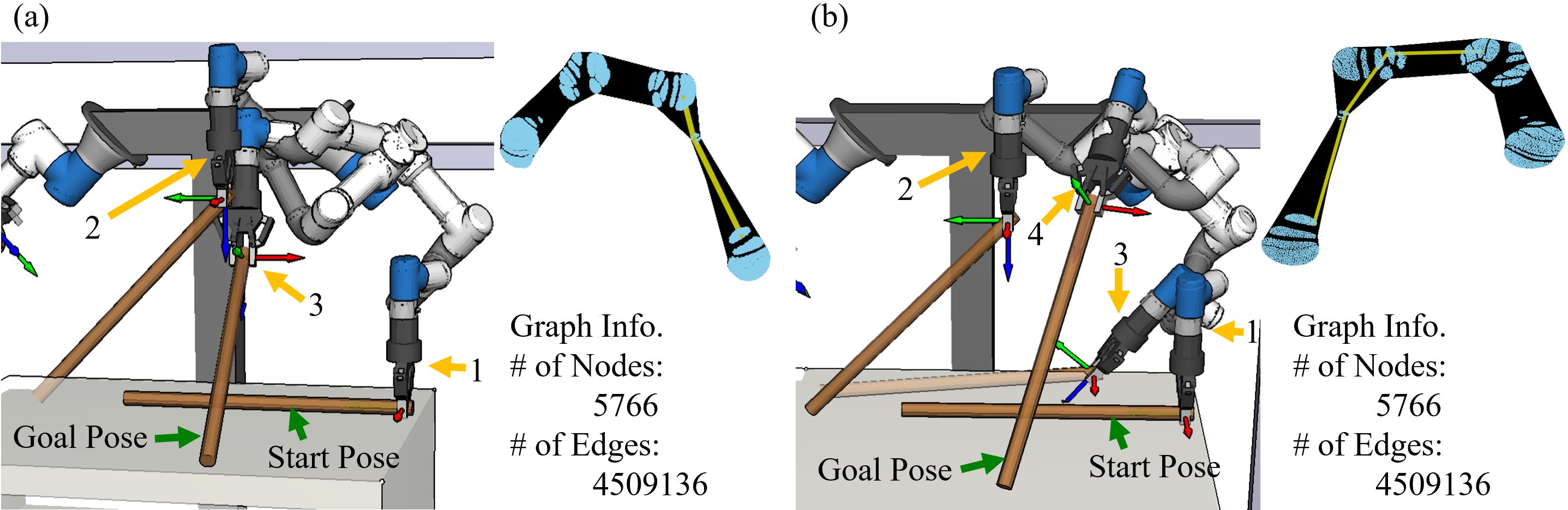}
    \caption{Results of the first experimental set. The goal of this set is to examine the drooping manipulation graph. (a) The key poses of the first task in this set. The sequence involves one grasp transition. The robot picks up the object at the start pose using grasp pose (1), moves it up to the transit pose using grasp pose (2) to change to a proper in-hand pose, and finally moves the object to the goal pose while keeping the same grasp pose. The right part of this subfigure shows the manipulation graph and the node/edge information. The yellow segments are the planned path. (b) Key poses of the second task. The sequence involves two grasp transitions. The robot moves up from grasp pose (1) to (2) to realize the first grasp transition. From grasp pose (2) to (3), the robot moves down while preserving the obtained grasp transition in the previous step. Then, from pose (3) to (4), the second grasp transition is conducted, and the object reached its goal pose. Like (a), the right part of this subfigure shows the manipulation graph and the node/edge information. The yellow segments are the planned path.}
     \label{fig:task1sim}
\end{figure}

\begin{figure}[!htb]
    \centering
\includegraphics[width=\textwidth]{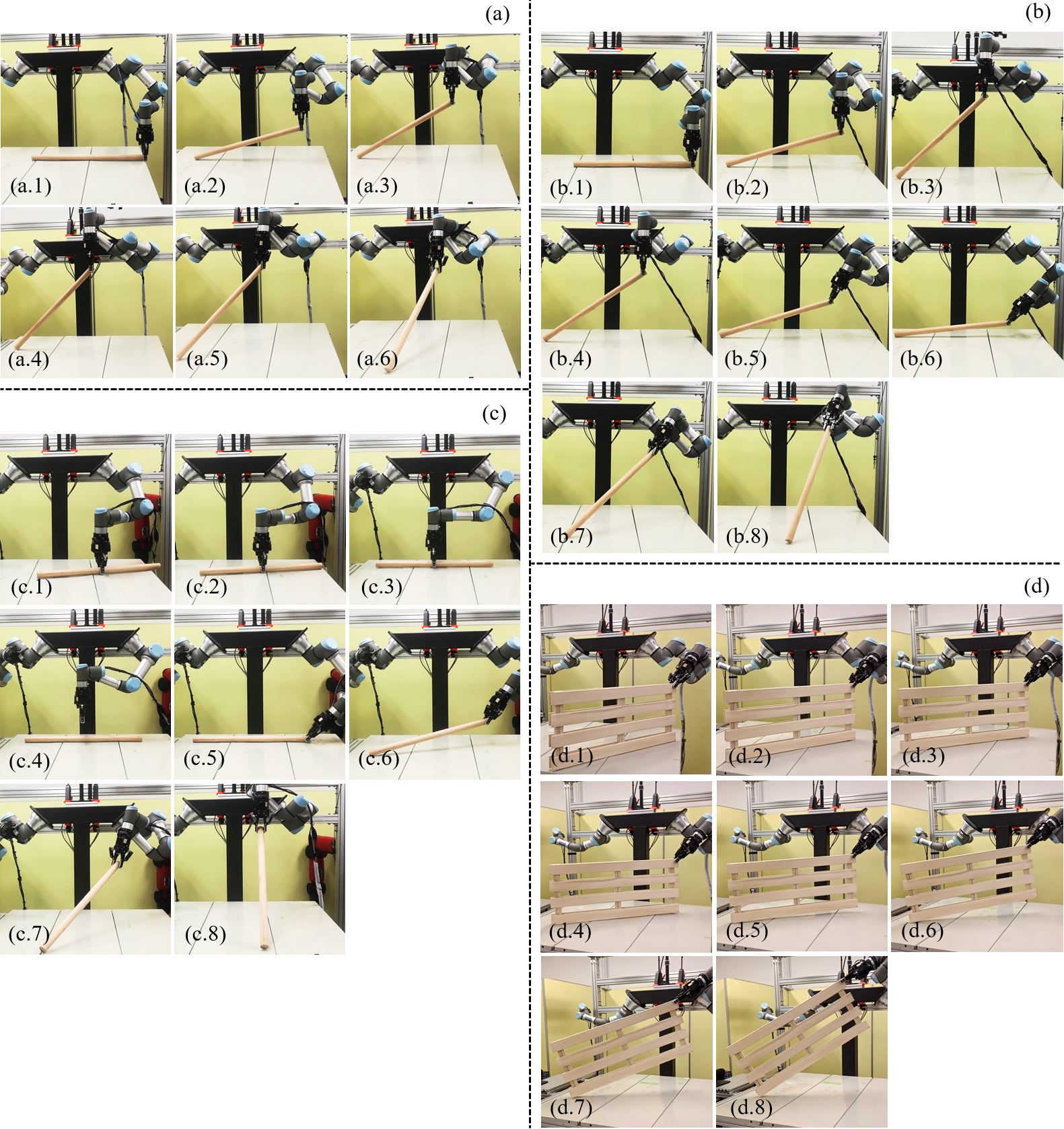}
        \caption{Real-world executions of the four tasks. (a-b) The two tasks in the first experiment set. (c-d) The two tasks in the second experiment set. Cartesian planning are used to interpolate the intermediate motion between the critical poses found from the manipulation graph.}
        \label{fig:results_subfig1}
\end{figure}

The first set contains two tasks. In the first task, we require the UR3 arm to move the wooden stick from a start pose on the table to a tilted goal pose facing the right direction. The start and goal poses are denoted by green arrows in Fig.\ref{fig:task1sim}(a). Using the drooping manipulation graph, the robot successfully found a sequence of critical poses to finish the task. The sequence is marked by ID (1)-(3) in the figure. It involves one time of in-hand grasp transition at the edge that connects ID (1) and (2). At edge (2)-(3), the robot kept the same grasp pose. The result of the real-world execution for this task is shown in Fig.\ref{fig:results_subfig1}(a). The second task's start and goal object poses are denoted by the green arrows in Fig.\ref{fig:task1sim}(b). The planner found a sequence involving two times of in-hand grasp transitions. The sequence of critical poses is denoted by ID (1)-(4) in the figure. The in-hand grasp transitions appeared at edges (1)-(2) and (3)-(4). The result of the real-world executions is shown in Fig.\ref{fig:results_subfig1}(b). The second sequence shows interesting behavior. The robot cannot complete the task by performing a direct upward motion because a configuration obstacle blocked the direct connection between the joint configurations at ID (1) and (4). To solve the problem, the planner tried transiting to the grasp pose at ID (2). The robot could move from ID (1) to (2) with a direct upward motion. However, the direct connection between ID (2) and (4) remained obstructed by configuration obstacles. The planner continued to search the graph and found an intermediate grasp pose ID (3). The robot may either directly move downward from ID (2) to (3) and move upward from (3) to (4), and thus could successfully finish the task. The critical grasp poses at ID (2) and (3) are the same in the object's local coordinate system. The edges at (1)-(2) and (3)-(4) indicate two in-hand grasp transitions.

\begin{figure}[!htb]
    \centering
    \includegraphics[width=\linewidth]{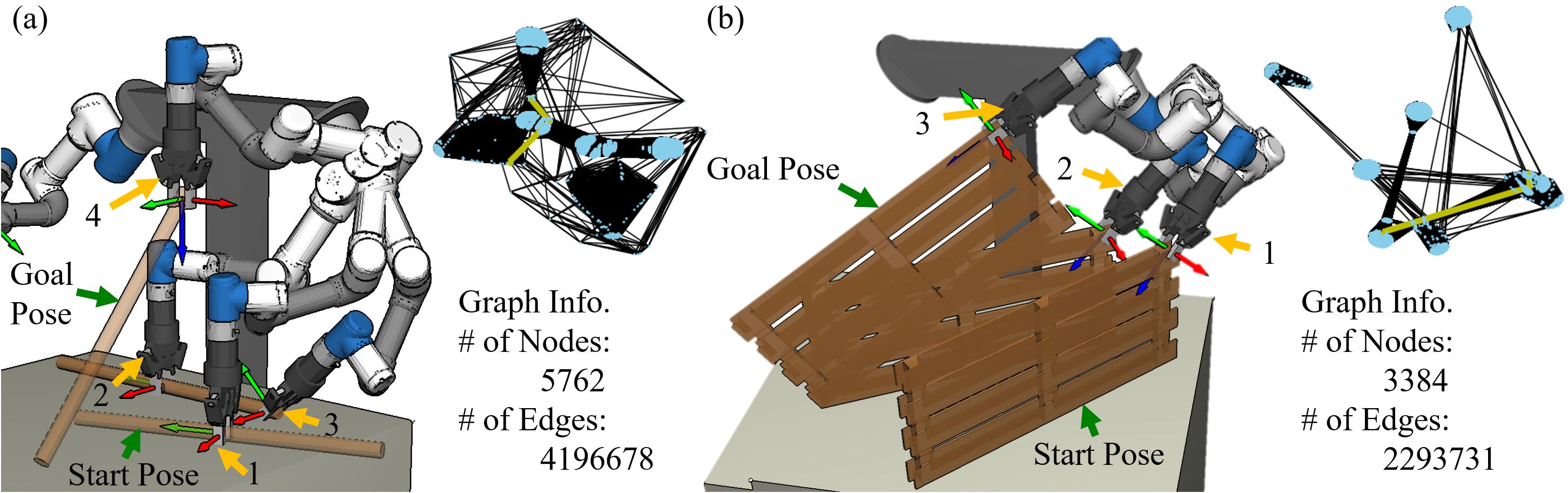}
    \caption{Results of the second experimental set. (a) The results of the first task in this set. The key poses include two regrasps and one transition. The robot grasps p the object using grasp pose (1) and slides it on the table to an intermediate pose while keeping the grasp. After that, the robot change grasp pose (3) to hold one end of the object. From pose (3) to (4), the object is delivered to its goal pose using constrained drooping. During the delivery, the grasp pose is transited to (4). (b) The planned sequence for the second task in this set. The planner finds a sequence where the robot slides the object to an intermediate pose for constrained drooping. Like Fig.\ref{fig:task1sim}, the right parts of the subfigures show the manipulation graph and the node/edge information. The yellow segments are the planned path. }
     \label{fig:task3sim}
\end{figure}

The second set of experiments aims to examine the planner's effectiveness in combining regrasping and constrained drooping. It also includes two tasks. In the first task, we ask the planner to move the stick from a start pose on the table from a goal pose facing forward, as is shown in Fig.\ref{fig:task3sim}(a). The path found by our planner involved both regrasping and constrained drooping motion. Since the start pose was far from the robot. The robot grasped the stick using grasp pose ID (1) and slided it to an intermediate pose. The grasp pose was kept the same in the process. Then, at the intermediate pose, the robot regrasped the object and changed its grasping pose to ID (3). Finally, the robot performed constrained droop to deliver the object to the goal pose. During the constrained drooping, the grasp poses were kept the same. There were no in-hand grasp transitions. The real-world executions of the planned sequence is shown in Fig.\ref{fig:results_subfig1}(a). In the second task, the robot is asked to manipulate a duck-bench. The sequence of key poses for this task is shown in Fig. \ref{fig:task3sim}(a). The planner found a sequence that brought the object to an intermediate pose for drooping. The intermediate pose was found at the connecting nodes in the combined graph, and thus the robot did not conduct release and regrasp. Instead, it directly delivered the object to its goal pose with the help of constrained drooping. The real-world executions of the sequence are shown in Fig.\ref{fig:results_subfig1}(b). The results showed that the combined planning of regrasping and constrained drooping effectively finds motion sequences for previously unsolvable tasks. Especially, regrasping enables a robot to repose an object to an appropriate state for drooping.

\section{Discussions}
\label{sec-dis}
The simulation and real-world results show the feasibility of our proposed method to manipulate long and heavy objects using a single arm and with the support of a table surface. The planner can autonomously decide regrasping and drooping intermediate object and grasp poses between the start and goal, and generate joint motion using Cartesian planning. Especially, the constrained drooping enabled changing the in-hand pose of the object and improved the connections among the key poses of the manipulated object.

While previous research has focused on using multiple robots to manipulate long and heavy objects, the results of this work demonstrate that a single arm with a supporting surface can be satisfactory for the same purpose. This finding can help reduce the scale of system integration because fewer manipulators are needed to solve the same problem. However, the generality of the results is subjected to limitations related to the objects' dimensions and physical properties. It is beyond this study's scope to decide the limit of physical properties of the objects that can be manipulated by a single arm and a support surface, and solve the related control and learning problems \cite{peng2020, huang2020, yang2019}. These out-of-scope topics are interesting directions for future studies. 

\section{Conclusions}
\label{sec-conclude}
We presented a planner for improving two-finger parallel grippers' dexterity to manipulate long and heavy objects. The planner could find robot motion sequences that manipulate objects while keeping them supported by an external support surface. The planner's essential part was the constrained drooping, which allowed tilting an object around a supporting point on the support surface with upward or downward motion. The planner considered a combination of constrained drooping and regrasping to build a manipulation graph and search the graph to get a manipulation sequence. The intermediate robot motions between the sequences were generated using Cartesian planning. The method was verified using various objects and tasks. The results showed that the method enabled using a single manipulator to maneuver long and heavy objects, rather than multiple arms as assumed in previous literature. In the future, we hope to refine the model of the soft-finger contact with tactile sensors and apply the method to objects with unknown materials and mass distributions.
\bibliographystyle{IEEEtran}
\bibliography{bibfile}

\end{document}